\documentclass[11pt]{article}
\pdfoutput=1
\usepackage{acl2016}
\usepackage[normalem]{ulem}
\usepackage{times}
\usepackage{url}
\usepackage{latexsym}
\usepackage{booktabs}
\usepackage{tabularx}
\usepackage{graphicx}
\usepackage{epstopdf}

\usepackage{array}
\usepackage[utf8]{inputenc}

\aclfinalcopy 
\def\aclpaperid{121} 

\usepackage{etoolbox}
\usepackage{multirow}
\usepackage{tocloft}
\usepackage[usenames,dvipsnames]{color}
\usepackage{changepage}

\newtoggle{showworknotes}
\toggletrue{showworknotes}

\newcommand{\listworknotestitle}{\large List of worknotes}
\newlistof{worknote}{wn}{\listworknotestitle}
\newcommand{\addworknote}[1]{%
\refstepcounter{worknote}%
\addcontentsline{wn}{worknote}%
{\protect\numberline{\theworknote}\ifstrempty{#1}{}{#1 }(\thesection)}}

\newcommand{\Prob}[1]{\ensuremath\mathsf{P}\left(#1\right)}

\def\perscite#1{\newcite{#1}}   
\def\parcite#1{\cite{#1}}    

\def\equo#1{``#1''}  
\def\samp#1{{\it #1}}  

\def\Sref#1{Section~\ref{#1}}
\def\Fref#1{Figure~\ref{#1}}
\def\Tref#1{Table~\ref{#1}}

\usepackage{glossaries}
\newacronym{nlp}{NLP}{natural language processing}
\newacronym{nn}{NN}{Neural Network}
\newacronym{rnn}{RNN}{Recurrent Neural Network}
\newacronym{cnn}{CNN}{Convolutional Neural Network}
\newacronym{wmt}{WMT}{Workshop of Machine Translation}
\newacronym{ape}{APE}{automatic post-editing}
\newacronym{smt}{SMT}{statistical machine translation}
\newacronym{mt}{MT}{machine translation}

\usepackage{breqn} 
\newcommand*{\ditto}{--- \raisebox{-0.5ex}{''} ---}

\title{CUNI System for WMT16 Automatic Post-Editing and Multimodal Translation Tasks} 

\author{Jindřich Libovický \qquad Jindřich Helcl \\ \bf Marek Tlustý \qquad Ondřej Bojar \qquad Pavel Pecina \\ \\
  Charles University in Prague \\
  Malostranské náměstí 25, 112~00 Prague, Czech Republic \\
  {\tt \{libovicky,helcl,tlusty,bojar,pecina\}@ufal.mff.cuni.cz}}

\date{}

\begin{document}
\maketitle
\begin{abstract}
Neural sequence to sequence learning recently became a very promising paradigm 
in machine translation, achieving competitive results with statistical phrase-based 
systems. In this system description paper, we attempt to utilize several recently published 
methods used for neural sequential learning in order to build 
systems for WMT 2016 shared tasks of Automatic Post-Editing
and Multimodal Machine Translation.
\end{abstract}

\section{Introduction}

Neural sequence to sequence models are currently used for variety of tasks in Natural Language
Processing including machine translation \parcite{sutskever2014sequence,bahdanau2014neural}, 
text summarization \parcite{rush2015summarization}, 
natural language generation \parcite{wen2015nlg},
and others. This was enabled by the capability of recurrent neural networks to model temporal 
structure in data, including the long-distance dependencies in case of gated networks 
\parcite{hochreiter1997lstm,cho2014gru}.

The deep learning models' ability to learn a dense representation 
of the input in the form of a real-valued vector
recently allowed researchers to combine machine vision and natural language processing into tasks 
believed to be extremely difficult only few years ago. The distributed representations of words, 
sentences and images can be understood as a kind of common data type for language and images within the models.
This is then used in tasks like automatic image captioning \parcite{vinyals2015show,xu2015show}, visual question 
answering \parcite{antol2015vqa} or in attempts to ground lexical semantics in vision \cite{kiela2015multi}.

In this system description paper, we bring a summary of the \gls{rnn}-based system
we have submitted to the automatic post-editing task and to the multimodal translation task.
\Sref{sec:description} describes the architecture of the networks we have used.
\Sref{sec:ape} summarizes related work 
on the task of automatic post-editing of machine translation output and describes our 
submission to the \gls{wmt} competition. In a similar fashion, \Sref{sec:mmmt} refers 
to the task of multimodal translation. Conclusions and ideas for further work are given in \Sref{sec:conclusion}.

\section{Model Description}
\label{sec:description}

\begin{figure*}
\centering
\includegraphics[scale=0.8]{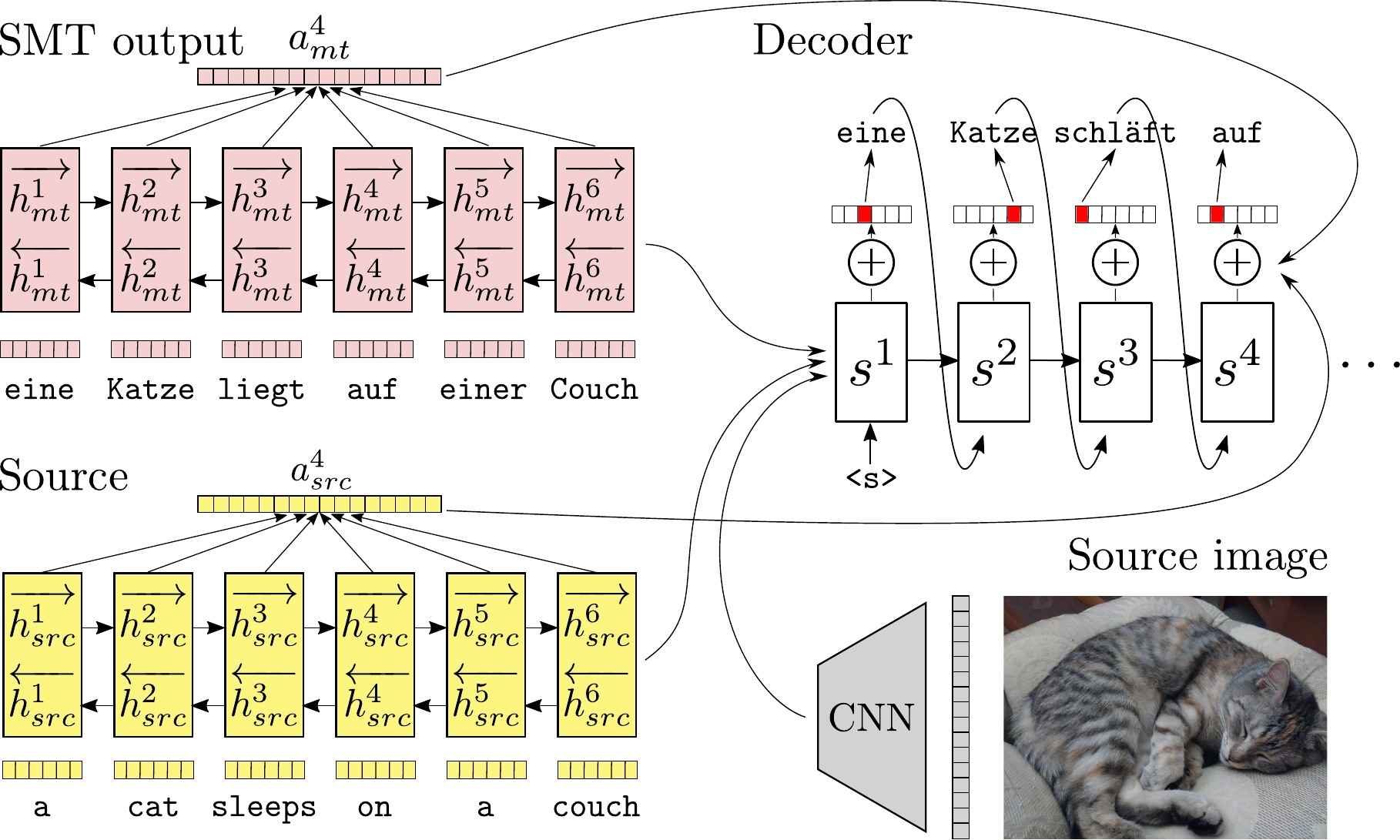}
\caption{Multi-encoder architecture used for the multimodal translation.}
\label{fig:arch}
\end{figure*}

We use the neural translation model with attention \parcite{bahdanau2014neural}
and extend it to include multiple encoders, see \Fref{fig:arch} for an illustration.
Each input sentence enters the system simultaneously in several representations $\mathbf{x_i}$.
An encoder used for the $i$-th representation
$\mathbf{X_i} = (x_i^1, \ldots, x_i^k)$
of $k$ words, each stored as a one-hot vector $x_i^j$,
is a bidirectional \gls{rnn}
implementing a function
\begin{equation}
f(\mathbf{X}_i) = \mathbf{H}_i = (h_i^1, \ldots, h_i^k)
\end{equation}
where the states $h_i^j$ are concatenations of the outputs of the forward and backward
networks after processing the $j$-th token in the respective order.

The initial state of the decoder is computed as 
a weighted combination of the encoders' final states.

The decoder is an \gls{rnn} which receives an embedding of the previously produced word as 
an input in every time step together with the hidden state from the previous time step. 
The \gls{rnn}'s output is then used to compute the attention 
and the next word distribution.

The attention is computed  over each encoder separately 
as described by \perscite{bahdanau2014neural}.
The attention vector $a_i^m$ of the $i$-th encoder in the $m$-th step of the decoder
is
\begin{equation}
a_i^m = \sum_{h_i^k \textnormal{ in } \mathbf{H}_i} h_i^k \alpha_i^{k,m}
 \end{equation}
where the weights $\alpha_i^{m}$ is a distribution estimated as
\begin{equation}
\alpha_i^m = \mathrm{softmax}\left(v^T \cdot \tanh (s^m + W_{\mathbf H_i} \mathbf{H}_i)\right)
\end{equation}
with $s^m$ being the hidden state of the decoder in time $m$. Vector $v$ 
and matrix $W_{\mathbf{H}_i}$ are learned parameters for projecting
the encoder states.

The probability of the decoder emitting the word $y_m$ 
in the $j$-th step, denoted as
$\Prob{y_m | \mathbf{H}_1, \ldots, \mathbf{H}_n, \mathbf{Y}_{0..m-1}}$,
is proportional to
\begin{dmath}
\exp \left( W_o s^j + \sum_{i=1}^n W_{a_i} a_i^j  \right)
\end{dmath}
where $\mathbf{H}_i$ are hidden states from the $i$-th encoder 
and $\mathbf{Y}_{0..m-1}$ is the already 
decoded target sentence (represented as matrix, one-hot vector for each produced word). 
Matrices $W_o$ and $W_{a_i}$ are learned parameters; 
$W_o$ determines the recurrent dependence on the decoder's state and $W_{a_i}$ 
determine the dependence on the (attention-weighted) encoders' states.

For image captioning, we do not use the attention model because of its high
computational demands and rely on the basic model by \perscite{vinyals2015show} instead.
We use Gated Recurrent Units \parcite{cho2014gru} and apply the dropout of 0.5 
on the inputs and the outputs of the recurrent layers \parcite{zaremba2014dropout}
and L2 regularization of $10^{-8}$ on all parameters. 
The decoding is done using a beam search of width 10. Both the decoders and
encoders have hidden states of 500 neurons, word embeddings have the dimension of 300.
The model is optimized using the Adam optimizer \parcite{kingma2015adam} with
learning rate of $10^{-3}$.

We experimented with recently published improvements of neural 
sequence to sequence learning: scheduled sampling \parcite{bengio2015scheduled}, 
noisy activation function \parcite{gulcehere2016noisy},
linguistic coverage model \parcite{zhaopeng2016coverage}. 
None of them were able to improve the systems' performance, so we do 
not include them in our submissions.

Since the target language for both the task was German, we also did language 
dependent pre- and post-processing of the text. For the training
we split the contracted prepositions and articles 
(\samp{am} $\leftrightarrow$ \samp{an dem}, 
\samp{zur} $\leftrightarrow$ \samp{zu der}, \ldots) 
and separated some pronouns from their case ending 
(\samp{keinem} $\leftrightarrow$ \samp{kein \hbox{-em}}, 
\samp{unserer} $\leftrightarrow$ \samp{unser \hbox{-er}}, \ldots).
We also tried splitting compound nouns into smaller units, but on the relatively
small data sets we have worked with, it did not bring any improvement.

\section{Automatic Post-Editing}
\label{sec:ape}

The task of \gls{ape} aims at improving the quality of a machine translation system treated as black box. 
The input of an \gls{ape} system is a pair of sentences -- the original input sentence in the source language 
and the translation generated by the \gls{mt} system. This scheme allows to use any 
\gls{mt} system without any prior knowledge of the system itself. 
The goal of this task is to perform automatic corrections on the translated sentences 
and generate a better translation (using the source sentence as an additional source of information).

For the \gls{ape} task, the organizers provided tokenized data from the IT domain \parcite{apedata2016}. 
The training data consist of 12,000 triplets of the source sentence, 
its automatic translation and a reference sentence. The reference sentences are 
manually post-edited automatic translations.
Additional 1,000 sentences were provided for validation, and another 2,000 sentences for final evaluation.
Throughout the paper, we report scores on the validation set; reference sentences 
for final evaluation were not released for obvious reasons.

The performance of the systems is measured using Translation Error Rate \parcite{snover:etal:2006}
from the manually post-edited sentences. We thus call the score HTER. 
This means that the goal of the task is more to simulate
manual post-editing, rather than to reconstruct the original unknown reference sentence.

\subsection{Related Work}

In the previous year's competition \cite{bojar2015wmt}, most of the systems were based on the phrase-base \gls{smt} in a monolingual setting \cite{simard2007postedit}.

There were also several rule-based post-editing systems benefiting 
from the fact that errors introduced by statistical and rule-based
systems are of a different type \parcite{rosa2015depfix,mohaghegh2013grafix}.
 
Although the use of neural sequential model is very straightforward in this case, 
to the best of our knowledge, there have not been experiments with \gls{rnn}s for this task.

\subsection{Experiments \& Results}
\label{ssec:ape_experiments}

The input sentence is fed to our system in a form of multiple input sequences without explicitly telling which sentence is the source one and which one is the MT output. 
It is up to the network to discover their best use when producing the (single) target sequence.
The initial experiments showed that the network struggles to learn
that one of the source sequences is almost correct (even if it shares the
vocabulary and word embeddings with the expected target sequence). Instead, the network seemed to learn to paraphrase the input.





To make the network focus more on editing of the source sentence instead of preserving the
meaning of the sentences, we represented the target sentence as a minimum-length sequence of edit
operations needed to turn the machine-translated sentence into the reference post-edit. We extended the vocabulary
by two special tokens \emph{keep} and \emph{delete} and then encoded the reference as 
a sequence of \emph{keep}, \emph{delete} and \emph{insert} operations with the insert operation
defined by the placing the word itself. See \Fref{tab:edits} for an example.

\def\linka#1{\textcolor{blue}{#1$_1$}}
\def\linkb#1{\textcolor{Maroon}{#1$_2$}}
\def\linkc#1{\textcolor{Green}{#1$_3$}}
\def\linkd#1{\textcolor{RedViolet}{#1$_4$}}
\def\linke#1{\textcolor{Bittersweet}{#1$_5$}}
\def\linkf#1{\textcolor{WildStrawberry}{#1$_6$}}
\def\linkg#1{\textcolor{MidnightBlue}{#1$_7$}}
\def\linkh#1{\textcolor{Red}{#1$_8$}}
\def\linki#1{\textcolor{OliveGreen}{#1$_9$}}
\def\linkj#1{\textcolor{Aquamarine}{#1$_{10}$}}
\def\linkk#1{\textcolor{Brown}{#1$_{11}$}}
\def\linkl#1{\textcolor{BurntOrange}{#1$_{12}$}}
\def\linkm#1{\textcolor{BrickRed}{#1$_{13}$}}
\def\linkn#1{\textcolor{BlueGreen}{#1$_{14}$}}
\def\linko#1{\textcolor{black}{#1$_{15}$}}

\begin{figure*}

\centering
\begin{tabular}{lp{12cm}}
\toprule
\bf Source & Choose Uncached Refresh from the Histogram panel menu. \\ \midrule
\bf MT & 
  \linka{Wählen} 
  \linkb{Sie}
  \linkc{Uncached}
  \linkd{"}
  \linke{Aktualisieren}
  \linkf{"}
  \linkg{aus}
  \linkh{dem}
  \linki{Menü} 
  \linkj{des}
  \linkk{Histogrammbedienfeldes}
  \linko{.} \\
\bf Reference & 
  \linka{Wählen} 
  \linkb{Sie} 
  \linkd{"} 
  \linkl{Nicht}
  \linkm{gespeicherte}
  \linkm{aktualisieren} 
  \linkf{"}
  \linkg{aus}
  \linkh{dem} 
  \linki{Menü} 
  \linkj{des}
  \linkk{Histogrammbedienfeldes}
  \linko{.} \\ \midrule
\bf Edit ops. &  
  \linka{\emph{keep}} 
  \linkb{\emph{keep}}
  \linkc{\emph{delete}} 
  \linkd{\emph{keep}} 
  \linkl{Nicht}
  \linkm{gespeicherte}
  \linkm{aktualisieren}
  \linke{\emph{delete}} 
  \linkf{\emph{keep}}
  \linkg{\emph{keep}}
  \linkh{\emph{keep}}
  \linki{\emph{keep}}
  \linkj{\emph{keep}}
  \linkk{\emph{keep}}
  \linko{\emph{keep}} \\
 \bottomrule
\end{tabular}

\caption{An example of the sequence of edit operations that our system should learn to produce when given the candidate MT translation. 
The colors and subscripts denote the alignment between the edit operations and the machine-translated
and post-edited sentence.}
\label{tab:edits}
\end{figure*}

\begin{table}[t]
\centering
\begin{tabular}{lcc}
\toprule
method                   & HTER     & BLEU  \\
\midrule
baseline                 &    .2481 &    62.29 \\


edit operations  &    .2438 & \bf 62.70 \\
edit operations+ & \bf.2436 &     62.62 \\
\bottomrule
\end{tabular}

\caption{Results of experiments on the APE task on the validation data. 
The `+' sign indicates the additional regular-expression rules -- the system
that has been submitted.}
\label{tab:postedit_val}

\end{table}




After applying the generated edit operations on the machine-translated sentences in the test phase,
we perform a few rule-based orthographic fixes for punctuation.
The performance of the system is given in \Tref{tab:postedit_val}.
The system was able to slightly improve upon the baseline (keeping the translation
as it is) in both the HTER and BLEU score. The system was able to deal very well 
with the frequent error of keeping a word from the source in the translated sentence.
Although neural sequential models usually learn the basic output structure very
quickly, in this case it made a lot of errors in pairing parentheses correctly.
We ascribe this to the edit-operation notation which obfuscated the basic orthographic 
patterns in the target sentences.

\section{Multimodal Translation}
\label{sec:mmmt}

The goal of the multimodal translation task is to generate an image caption
in a target language (German) given the image itself and one or more
captions in the source language (English).

Recent experiments of \perscite{elliott2015multilanguage} showed that including the 
information from the images can help disambiguate the source-language
captions.

The participants were provided with the Multi30k dataset \parcite{elliot2016multi30k} 
which is an extension of the Flickr30k dataset \parcite{plummer2015flickr30k}.
In the original dataset, 31,014 images were taken from the users collections
on the image hosting service Flickr. Each of the images were given five independent
crowd-sourced captions in English. For the Multi30k dataset, one of the English
captions for each image was translated into German and five other independent
German captions were provided. The data are split into a training set of 29,000
images, a validation set of 1,014 images and a test set with 1,000 images.

The two ways in which the image annotation were collected also lead to two sub-tasks. 
The first one is called Multimodal Translation and its goal is to generate
a translation of an image caption to the target language given the caption in
source language and the image itself. The second task is the Cross-Lingual Image Captioning.
In this setting, the system is provided five captions in the source language
and it should generate one caption in target language given both source-language captions
and the image itself. Both tasks are evaluated using the BLEU \parcite{papineni2002bleu}
score and METEOR score \parcite{denkowski2011meteor}.
The translation task is evaluated against a single reference sentence which is the
direct human translation of the source sentence. The cross-lingual captioning task
is evaluated against the five reference captions in the target language created
independently of the source captions.

\subsection{Related Work}

\begin{table*}
\centering
\begin{tabular}{lcccc}
\toprule
            & \multicolumn{2}{c}{Multimodal translation} & \multicolumn{2}{c}{Cross-lingual captioning} \\
 
 system                            & BLEU     & METEOR   & BLEU     & METEOR   \\ 
 \midrule
  Moses baseline                   &     32.2 &     54.4 &     11.3 &     33.8 \\
 MM baseline                       &          &     27.2 &          &     32.6 \\ 
 
 \midrule
 
 tuned Moses                       &     36.8 & \bf 57.4 &     12.3 &     35.0 \\
 NMT                               &     37.1 &     54.6 &     13.6 &     34.6 \\
 NMT + Moses                       &     36.5 &     54.3 &     13.7 &     35.1 \\
 NMT + image                       &     34.0 &     51.6 &     13.3 &     34.4 \\
 NMT + Moses + image               & \bf 37.3 &     55.2 &     13.6 &     34.9 \\ 
 \ditto, submitted                 &     31.9 &     49.6 &     13.0 &     33.5 \\
 
 \midrule
 
 captioning only                   &          &          &      9.1 &     25.3 \\
 5 en captions                     &          &          &     22.7 &     38.5 \\
 5 en captions + image             &          &          & \bf 24.6 & \bf 39.3 \\
 \ditto, submitted                 &          &          &     14.0 &     31.6 \\
 \bottomrule
\end{tabular}

\caption{Results of experiments with the multimodal translation task on the validation data. 
At the time of the submission, the models were not tuned as well as our final models. The first
six system are targeted for the translation task. They were trained against one reference -- 
a German translation of one English caption. The last four systems are target to 
the cross-lingual captioning task. They were trained with 5 independent German captions (5 times bigger data).}

\label{tab:mmmt_task}
\end{table*}

The state-of-the-art image caption generators use a remarkable property of the \gls{cnn} models
originally designed for ImageNet classification to capture the semantic features
of the images. Although the images in ImageNet \parcite{deng2009imagenet,russakovsky2015imagenet}
always contain a single
object to classify, the networks manage to learn a representation that is usable
in many other cases including image captioning which usually concerns multiple
objects in the image and also needs to describe complex actions and spacial and temporal relations within the image.

Prior to \gls{cnn} models, image classification
used to be based on finding some visual primitives in the image
and transcribing automatically estimated relations between the primitives.
Soon after \perscite{kiros2014multimodal} showed that the \gls{cnn} features could be used
in a neural language model, \perscite{vinyals2015show} developed a model that used an \gls{rnn}
decoder known from neural \gls{mt} for generating captions from the image features
instead of the vector encoding the source sentence. \perscite{xu2015show} later even improved
the model by adapting the soft alignment model \parcite{bahdanau2014neural} nowadays 
known as the attention model. Since then, these models have become a benchmark for works trying
to improve neural sequence to sequence models 
\parcite{bengio2015scheduled,gulcehere2016noisy,ranzato2015mixer}.

\subsection{Phrase-Based System}

For the translation task, we trained Moses \gls{smt} \parcite{koehn-EtAl:2007:PosterDemo}
with additional language models based on coarse bitoken classes. 
We follow the approach of \perscite{stewart:etal:2014}: Based on the word alignment, each target
word is concatenated with its aligned source word into one bitoken (e.g.\equo{Katze-cat}).
For unaligned target words, we create a bitoken with NULL as the source word
(e.g. \equo{wird-NULL}). Unaligned source words are dropped. For more than one-to-one alignments, we join all aligned word pairs
into one bitoken (e.g. \equo{hat-had+gehabt-had}). These word-level bitokens are
afterwards clustered
into coarse classes \parcite{brown:etal:1992} and a standard $n$-gram 
language model is trained on these classes. Following the notation 
of \perscite{stewart:etal:2014}, \equo{400bi} indicates a LM trained on 400 bitoken classes, \equo{200bi} stands for 200
bitoken classes, etc.
Besides bitokens based on aligned words, we also use class-level bitokens. For example \equo{(200,400)}
means that we clustered source words into 200 classes and target words into 400
classes and only then used the alignment to extract bitokens of these coarser words.
The last type is \equo{100bi(200,400)}, a combination of both independent 
clustering in the source and target \equo{(200,400)} and the bitoken clustering \equo{100bi}.

Altogether, we tried 26 configurations combining various coarse language
models. The best three were \equo{200bi} (a single bitoken LM), 
\equo{200bi\&(1600,200)\&100tgt} (three LMs, each with its own weight, 
where 100tgt means a language model over 100 word classes trained on the target side only) and
\equo{200bi\&100tgt}.

Manual inspection of these three best configurations reveals almost no differences; often the outputs are
identical. Comparing to the baseline (a single word-based LM), it is evident that coarse models prefer to ensure
agreement and are much more likely to allow for a different word or preposition choice to satisfy the agreement.


\subsection{Neural System}
\def\err#1{\uline{#1}}
\def\strange#1{\uwave{#1}} 

\begin{figure*}
\centering
\scalebox{0.87}{
\begin{tabular}{lp{0.55\textwidth}c}
\toprule

\textbf{Source} 
  & A group of men are loading cotton onto a truck
  & \multirow{7}{*}{\includegraphics[width=0.3\textwidth]{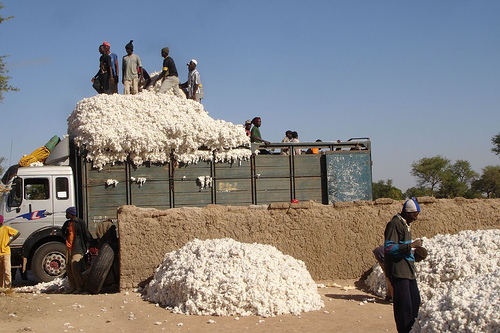}} \\
\textbf{Reference} & Eine Gruppe von Männern lädt Baumwolle auf einen Lastwagen \\
\textbf{Moses}  & eine Gruppe von Männern lädt \err{cotton} auf einen \err{Lkw} \\
\quad\textit{2 Errors:}& \textit{untranslated \equo{cotton} and capitalization of \equo{LKW}} \\ 
\textbf{MMMT} & Eine Gruppe von Männern lädt \strange{etwas} auf einem \err{Lkw}. \\
\quad\textit{Gloss:}& \textit{A group of men are loading \strange{something} onto a truck.} \\
\textbf{CLC} & Mehrere Personen stehen an einem LKW. \\
\quad\textit{Gloss:}& \textit{More persons stand on a truck.} \\

\midrule 

\textbf{Source}
  & A man sleeping in a green room on a couch.
  &  \multirow{6}{*}{\includegraphics[width=0.3\textwidth]{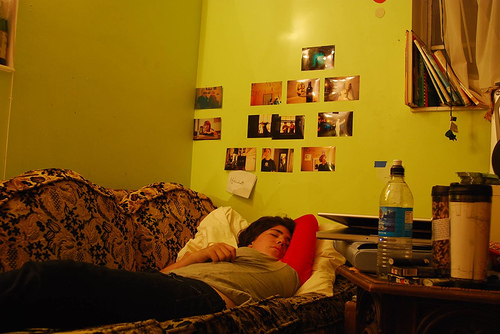}}\\
\textbf{Reference} & Ein Mann schläft in einem grünen Raum auf einem Sofa. \\
\textbf{Moses} & Ein Mann schläft in einem grünen Raum auf einem Sofa. \\
\textbf{MMMT} & Ein Mann schläft in einem grünen Raum auf einer Couch. \\
& \textit{No error, a correctly used synonym for \equo{couch}.} \\
\textbf{CLC} & Eine Frau schläft auf einer Couch. \\
& \textit{A man (\equo{Mann}) is mistaken for a woman (\equo{Frau}).} \\

\bottomrule
\end{tabular}}
\caption{Sample outputs of our multimodal translation (MMMT) system and cross-lingual captioning (CLC) system 
in comparison with phrase-based MT and the reference. The \emph{MMMT} system refers to the `NMT + Moses + image' row
and \emph{CLC} system to the `5 captions + image' row in \Tref{tab:mmmt_task}.}
\label{tab:mmmt_output}
\end{figure*}

For the multimodal translation task, we combine the \gls{rnn} encoders with image features.
The image features are extracted from the 4096-dimensional penultimate layer (\emph{fc7}) 
of the VGG-16 Imagenet network \cite{simonyan2014vgg} before applying non-linearity. 
We keep the weights of the convolutional network
fixed during the training. We do not use attention over the image features, so the 
image information is fed to the network only via the initial state.

We also try a system combination and add an encoder for the phrase-based output. The \gls{smt}
encoder shares the vocabulary and word embeddings with the decoder. For the combination with 
\gls{smt} output, we experimented with the CopyNet architecture \parcite{jiatao2016copynet}
and with encoding the sequence the way as in the \gls{ape} task 
(see \Sref{ssec:ape_experiments}). Since neither of these variations seems to have any effect on the
performance, we report only the results of the simple encoder combination.

Systems targeted for the multimodal translation task have a single English caption 
(and eventually its SMT and the image representation) on its input and produce
a single sentence which is a translation of the original caption. 
Every input appears exactly once in the training data
paired with exactly one target sentence. On the other hand, systems targeted for the cross-lingual captioning
use all five reference sentences as a target, i.e. every input is present five times in the training data
with five different target sentences, which are all independent captions in German.
In case of the cross-lingual captioning, we use five parallel encoders sharing all weights
combined with the image features in the initial state.

Results of the experiments with different input combinations are summarized in the next section.

\subsection{Results}

The results of both the tasks are given in \Tref{tab:mmmt_task}. 
Our system significantly improved since the competition submission, therefore we report both the
performance of the current system and of the submitted systems. Examples of the
system output can be found in \Fref{tab:mmmt_output}.

The best performance has been achieved by the neural system that combined all available input both 
for the multimodal translation and cross-lingual captioning.
Although, using the image as the only source of information led to poor results, adding 
the image information helped to improve the performance in both tasks. 
This supports the hypothesis that for the translation of an image caption, knowing the
image can add substantial piece of information.

The system for cross-lingual captioning tended to generate very short descriptions, which were usually
true statements about the images, but the sentences were often too general or missing important information.
We also needed to truncate the vocabulary which brought out-of-vocabulary tokens to the system output.
Unlike the translation task where the vocabulary size was around 20,000
different forms for both languages, having 5 source and 5 reference sentences increased
the vocabulary size more than twice.

Similarly to the automatic postediting task, we were not able
to come up with a setting where the combination with the phrase-based system
would improve over the very strong Moses system with bitoken-classes language model.
We can therefore hypothesize that the weakest point of the models
is the weighted combination of the inputs for the initial state of the decoder. The difficulty of 
learning relatively big combination weighting matrices which are used just once during the model 
execution (unlike the recurrent connections having approximately the same number of parameters) probably 
over-weighted the benefits of having more information on the input. In case of system combination, 
more careful exploration of explicit copy mechanism as CopyNet \parcite{jiatao2016copynet} may be useful.

\section{Conclusion}
\label{sec:conclusion}

We applied state-of-the art neural machine translation models to
two \gls{wmt} shared tasks. We showed that neural sequential models could be successfully applied
to the \gls{ape} task. We also showed that information from the image can significantly help while
producing a translation of an image caption. Still, with the limited amount of data provided,
the neural system performed comparably to a very well tuned \gls{smt} system.

There is still a big room for improvement of the performance
using model ensembles or recently introduced techniques for neural sequence to sequence learning. 
An extensive hyper-parameter testing could be also helpful.

\section*{Acknowledgment}

We would like to thank Tomáš Musil, Milan Straka and Ondřej Dušek for discussing
the problem with us and countless tips they gave us during our work.

This work has received funding from the European Union's Horizon 2020 research
and innovation programme under grant agreement
no. 645 452 (QT21) and no. 644 753 (KConnect) 
and the Czech Science Foundation (grant n. P103/12/G084).
Computational resources were provided by the CESNET LM2015042, provided under the programme ``Projects of Large Research, Development, and Innovations Infrastructures.''


\bibliography{mmmt}
\bibliographystyle{acl2016}



\end{document}